\documentclass[a4paper]{article}
\usepackage{algorithmic}
\usepackage{algorithm}
\usepackage{multirow}
\usepackage{INTERSPEECH2020}
\title{Is Everything Fine, Grandma? Acoustic and Linguistic Modeling for Robust Elderly Speech Emotion Recognition}

\name{Gizem So\u{g}anc{\i}o\u{g}lu$^1$, Oxana Verkholyak$^{2}$, Heysem Kaya$^1$, Dmitrii Fedotov$^{3,4}$, Tobias Cad\'ee$^1$, \\Albert Ali Salah$^1$,  Alexey Karpov$^2$}
\address{
$^1$ Department of Information and Computing Sciences, Utrecht University, Utrecht, The Netherlands
$^2$ St. Petersburg Institute for Informatics and Automation of the Russian Academy of Sciences \\ SPIIRAS, St. Petersburg, Russia\\
$^3$ Ulm University, Ulm, Germany\\
$^4$ ITMO University, St. Petersburg, Russia
 }
\email{
g.sogancioglu@uu.nl,
overkholyak@gmail.com, 
h.kaya@uu.nl,
dmitrii.fedotov@uni-ulm.de,
cadeetobias@gmail.com,
a.a.salah@uu.nl,
karpov@iias.spb.su
}

\begin{document}

\maketitle
\begin{abstract}
Acoustic and linguistic analysis for elderly emotion recognition is an under-studied and challenging research direction, but essential for the creation of digital assistants for the elderly, as well as unobtrusive telemonitoring of elderly in their residences for mental healthcare purposes. This paper presents our contribution to the INTERSPEECH 2020 Computational Paralinguistics Challenge (ComParE) - Elderly Emotion Sub-Challenge, which is comprised of two ternary classification tasks for arousal and valence recognition. We propose a bi-modal framework, where these tasks are modeled using state-of-the-art acoustic and linguistic features, respectively.
In this study, we demonstrate that exploiting task-specific dictionaries and resources can boost the performance of linguistic models, when the amount of labeled data is small. Observing a high mismatch between development and test set performances of various models, we also propose alternative training and decision fusion strategies to better estimate and improve the generalization performance.

\end{abstract}


\section{Introduction}
\label{sec:intro}
While the state-of-the-art in affective computing and paralinguistic analysis reaches new peaks, research on two subject groups, namely the children and the elderly, lags behind due to scarcity of training resources and difficulty of data collection~\cite{wang2013database,KAYACSL17}. Acoustic characteristics of these groups differ significantly from other age groups mainly found in the available datasets. Models trained on available data do not perform optimally for recognition in extreme age groups. Computational Paralinguistics Challenge (ComParE) 2020~\cite{compare2020, markitantov2020mask} introduces a novel elderly emotion dataset, where both acoustic signals and speech transcriptions are provided.

Recent ComParE challenges introduced new feature types, such as Bag-of-Audio-Words~\cite{schmitt2017openxbow} and embeddings from Sequence-to-Sequence Deep Recurrent Neural Networks ~\cite{freitag2017audeep}. The results of these challenges showed that ensemble systems and alternative feature representations have a great potential for advancing the state-of-the-art. However, the constituents and the combination rules of the ensemble systems must be selected with care. Kaya and colleagues previously applied the Fisher Vector (FV) encoding of acoustic Low-Level Descriptors (LLD) to several paralinguistic tasks including recognition of native language and sincerity~\cite{kaya2016fusing}, as well as classification of snoring types~\cite{kaya2017introducing} and eating conditions~\cite{kaya2015fisher}. In this work, we use a similar FV encoding for the representation of acoustic features.

In affective computing and paralinguistics research, it is known that acoustic models perform well for arousal recognition, while providing poorer performance on valence recognition~\cite{SchullerCAHB, KayaetalAVEC19, ElAyadi2011ReviewSER}. Leveraging video and spoken content, when they are available, provides significant improvement on both valence and categorical emotion recognition performance~\cite{ElAyadi2011ReviewSER}. 
Based on our experiences and the relevant literature~\cite{compare2020, SchullerCAHB, KayaetalAVEC19}, we propose to leverage the linguistic modality for valence and the acoustic modality for arousal in this work. We propose modality-specific ensemble systems for arousal and valence recognition, while investigating the effectiveness of acoustic and linguistic models on both recognition tasks to support our hypothesis. For robust valence modeling, we extract a set of state-of-the-art linguistic features, including TF-IDF (Term Frequency-Inverse Document Frequency), FastText word embeddings, high-level polarity features, and dictionary-based linguistic features in German and English.

The contribution of this work is manifold. Firstly, we propose a bi-modal framework leveraging the linguistic modality for valence and the acoustic modality for arousal prediction. Secondly, we extract and experiment with a plethora of state-of-the-art acoustic and linguistic feature sets. Thirdly, we investigate strategies for fusion (at both feature and decision levels) and modeling with a high generalization power. We apply the proposed systems and strategies on the ComParE-2020 Elderly Emotion Sub-Challenge and obtain a marked improvement over the challenge test set baselines. 
\section{Background on Methods}
\label{sec:bg}
Our task in this work is to predict valence and arousal ratings (in Low, Medium, High levels) of spontaneous narratives (i.e. acoustic and linguistic modalities). The reader is referred to~\cite{compare2020} for details of the baseline acoustic and linguistic features, as well as the challenge corpora. Here, we briefly provide the background on methods we used from the literature. 

\subsection{Fisher Vector Representation for Acoustic Descriptors}
\label{subsec:fv}
The Fisher Vector (FV) encoding ~\cite{perronnin2007fisher} is a state-of-the-art  representation method for representing low level descriptors (LLD) over an image, utterance or video, firstly introduced in the computer vision domain, and successfully applied in paralinguistic analysis ~\cite{kaya2016fusing, kaya2017introducing}. For the emotion estimation task, we train FVs for the acoustic modality at utterance level. FV requires a background probability model, typically a Gaussian Mixture Model (GMM), trained on LLDs of all the utterances (more on this later) from the training set. 
Normally, this requires the computation of the Fisher information matrix, which can be approximated in the case when diagonal covariance matrices are used with the GMM. The set of LLDs are de-correlated and projected to a lower dimension using Principal Component Analysis (PCA). Hence, the number of PCA components $K_{PCA}$ and GMM clusters $K_{GMM}$ are the hyper-parameters of the FV encoding.

\subsection{Sentiment Dictionaries}
\label{subsec:dictionaries}
We use two sentiment dictionaries to estimate emotion from language use. These are language specific resources that contain a list of affective words with associated positive or negative scores. The SentiWS dictionary~\cite{remus2010sentiws} contains 3467 German words together with the corresponding inflections and Part-of-Speech (POS) tags. Each word is assigned a single score, which is estimated using frequencies and co-occurrence statistics on a German-language corpus consisting of approximately 100M sentences. The scores are scaled to the range of [-1.0, 1.0] with +1.0 being absolutely positive and -1.0 being absolutely negative. The SentiWordNet dictionary~\cite{baccianella2010sentiwordnet} is based on WordNet and consists of almost 207K English words (60 times more than SentiWS) together with POS tags. Each word is assigned a positive and a negative score in the range of [0.0, 1.0]. 

\subsection{Supervised Classifiers}
\label{subsec:classifiers}
Due to their popularity in handling high-dimensional feature vectors (e.g. supra-segmental acoustic features), Support Vector Machines (SVMs) are used in many emotion classification systems. In our ensemble, we additionally employ Kernel Extreme Learning Machines (ELM)~\cite{huang2012extreme} and Partial Least Squares (PLS) regression~\cite{wold1985partial}, since these are fast and accurate algorithms that previously produced state-of-the-art results on several speech-based and multimodal tasks~\cite{kaya2016fusing, kaya2017introducing}. We obtain kernels from the training data for both PLS and ELM. For handling data imbalance, we employ a variant of ELM dubbed \textit{Weighted Kernel ELM (WKELM)}~\cite{Zong2013229} (and its Kernel PLS counterpart WKPLS that we introduced in~\cite{kaya2017introducing}), which assign higher weights to minority class instances during model learning.

In addition to kernel-based classifiers, we employ Gradient Boosting Machines (GBM) in this work. GBMs are a special family of decision tree ensembles, where the $K^{th}$ tree is trained to predict the residual from the former $K-1$ trees~\cite{friedman2001greedy,chen2016xgboost}. In GBMs, tree learning is boosted with instance-wise gradient and the Hessian of the loss function.


\section{System Development}
\label{sec:appro}
\begin{figure}[htbp]
	\centering
	\includegraphics[width=\columnwidth]{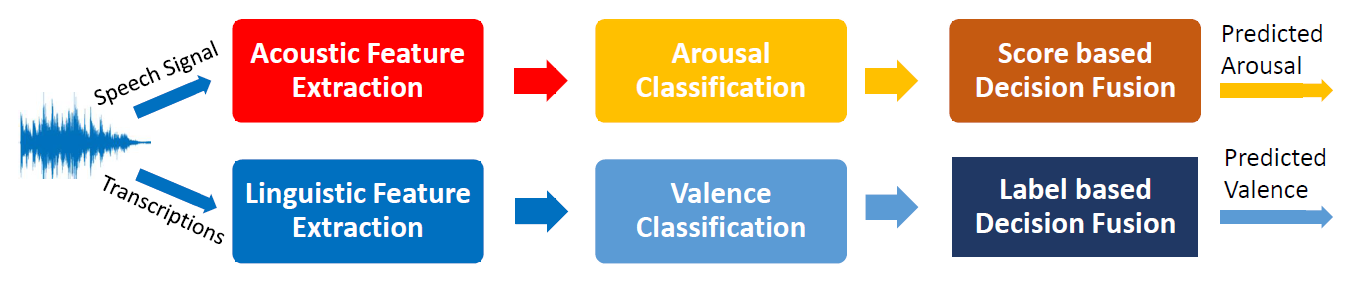}
	\caption{The proposed pipeline for bi-modal elderly speech emotion recognition.}
	\label{fig:pp_overall}
\end{figure}

An overview of the proposed bi-modal arousal and valence recognition system is given in Figure~\ref{fig:pp_overall}. The main idea is to leverage the strength of different modalities in each task. The details of feature extraction, classification and fusion steps are given in the following subsections.

\subsection{Acoustic Feature Extraction}
We extract FV based features that were shown to be effective in former paralinguistic challenges~\cite{kaya2016fusing,kaya2017introducing}. As acoustic LLDs, we extract Mel-Frequency Cepstral Coefficients (MFCCs) 0-24 and RASTA-PLP (Perceptual Linear Prediction) cepstrum for 12$^{th}$ order linear prediction, together with their temporal $\Delta$ coefficients, making an LLD vector of 76 dimensions. The combined LLDs are PCA-projected to preserve 99.9\% of the variation in the data. This keeps the dimensionality virtually unchanged (75 dimensions), but de-correlates the data. The number of GMM clusters are optimized using cross-validation (CV).

\subsection{Linguistic Feature Extraction}
\label{subsec:ling}
Transformer language embeddings (such as BERT~\cite{devlin2018bert}) are the state-of-the-art in representing linguistic features, and are also a part of the baseline system for the given sub-challenge. However, as seen from the baseline performance, they may show unreliable results on small datasets. Therefore, we test four alternative representations: FastText, dictionary-based, high-level polarity, and TF-IDF features, respectively. We explain these features in dedicated subsections. 

In addition to representations obtained from the original German transcriptions, we propose to extract features from the automatic English translations obtained via Google Translation engine\footnote{https://cloud.google.com/translate}. Analysing English texts is advantageous, since there are more sentiment analysis resources available for English.

\subsubsection{FastText Embeddings}
We use the FastText model~\cite{bojanowski2017enriching}, which is a state-of-the-art approach for character level embeddings, producing a semantic vector representation of words in a story. We use pre-trained 100-dimensional English and German word embeddings~\cite{grave2018learning}, which are trained on Common Crawl\footnote{http://commoncrawl.org/}, and finetune the pre-trained model on our dataset. The story-level vectors are obtained by averaging the 100-dimensional vectors of each word that construct the corresponding story. 

\subsubsection{Dictionary-based Features}
In order to obtain story-level SentiWS scores, an input text is tokenized ignoring the punctuation, and each token is looked up in the dictionary. If it is not found, the list of inflections for each word in the dictionary is checked without POS tag. Usually this results in a single score, but if multiple matches are found, a mean between the scores is accepted as a final score. The output of this process is a sequence of scores for tokens found in the dictionary. The following statistics are applied to the sequence: minimum, maximum, range, mean, sum, and numbers of positive and negative scores. 

For the SentiWordNet representation, an input text is tokenized ignoring the punctuation, and each token is looked up according to its POS. It is common to see multiple matches since English is rich in homographs and SentiWordNet disambiguates between many of them. The mean score is used as a final score. If a token is not found in the dictionary, the same process is repeated for its lemma using original token's POS. The outputs of this process are two sequences containing positive and negative scores for each token found in the dictionary. Then the sequences are used to calculate two sets of the following statistics: minimum, maximum, range, mean, sum and number of instances. The tokenization, POS tagging and lemmatization were performed using the NLTK Python library~\cite{bird2009natural}. 

\subsubsection{High-level Polarity Features}
As polarity and subjectivity features, we use available sentiment analysis tools, namely NLTK Vader~\cite{hutto2014vader}, TextBlob~\cite{loria2018textblob} and Flair~\cite{akbik2019flair}. Each of these libraries have some strengths and drawbacks in assessing the sentiment of the sentences. For instance, TextBlob is based on a simple pattern analyzer logic and fails to take negation into account in the sentence. However, alongside polarity prediction, it applies subjectivity analysis, which can be considered as a good feature for the valence dimension. Vader is good at handling negation thanks to some heuristics, but performs weakly on unseen words. Flair, which is based on a character-level Long Short-Term Memory (LSTM) network, is good at assessing the polarity of unseen words (such as those that result from typos).

To benefit from the strengths of each approach, we use the predicted polarity and subjectivity probability scores of each library as high-level features for our model. Since English is the common supported language among these tools, we use English transcripts (machine translated from German) to extract these features. Although these methods are designed to work on sentence-level or shorter length of text, we applied them at the story-level yet still obtain a good performance.

\subsubsection{TF-IDF Model}
We extract Term Frequency-Inverse Document Frequency (TF-IDF) representation as an additional linguistic feature. This representation is commonly used in natural language processing~\cite{tfidf2014knn}, information retrieval~\cite{tfidf_ir} and text mining~\cite{tfidf_text_mining} tasks. As a vital step of using lexical level features, we apply standard pre-processing methods, such as removal of stop words by using the English/German stop words dictionaries available in the NLTK library
~\cite{bird2009natural}, and stemming by the Porter stemmer~\cite{porter1980algorithm}. Afterwards, TF-IDF weights are computed over the set of uni-grams and bi-grams. 

\subsection{Proposed Model Training and Optimization Strategy}
\label{subsec:proptim}
A typical training strategy under the challenge protocol is as follows. We train the models on the challenge \textit{training set}, optimize the hyper-parameters and the feature types on the \textit{development set}. Once an optimal setting is found, we combine both and re-train with the optimal setting. Generalization to the sequestered test set depends on many factors, and is difficult to predict.
To overcome this problem, we propose to use N-Fold cross-validation (CV) to generate N learners and to fuse their decisions for the test set. This effectively increases the training data, shows the model's performance on the entire set of annotated data points and also reduces the test set error via combining multiple learners. To estimate the real-life (in our case challenge test set) performance, we propose to use a nested N-Fold CV. 

\subsection{Proposed Label Fusion Strategies}
\label{subsec:labfusion}
For fusing class labels, it is common to use majority voting of three or more models. We extend this approach, mainly for breaking ties, and we propose a rule based strategy for fusing two prediction sets. The proposed tie break mechanisms benefit from our domain knowledge, namely the ordinal structure of the target variables and the class distribution. 

The tie break mechanism for three models in a ternary classification uses an idea that in such a tie all three levels are predicted by base models, thus finds the middle way, outputting the medium class. In the case a higher number ($>$ 3) of models are combined, it favors the minority class(es). The tie break mechanism for two models further considers the development set confusion matrices to infer the bias towards majority and minority classes of the two models. A rule set used for combining a prediction set favoring the majority class ($P$) and a prediction set favoring the minority classes ($S$) is given in Algorithm~\ref{alg:algorithm_fuse2}.

\begin{algorithm}
    \caption{Fusion of two prediction sets \{$P_i$, $S_i$\} into \{$O_i$\}, i=1\ldots N, where the class labels \{`L', `M', `H'\} are ordinal.}
    \label{alg:algorithm_fuse2}
    \begin{algorithmic}
        \IF{$P_i$ = $S_i$}
        \STATE $O_i$ $\leftarrow$ $P_i$
    \ELSIF{$P_i$ and $S_i$ contain opposite extreme labels}
        \STATE $O_i$ $\leftarrow$ `M'
    \ELSIF{$S_i$ contains a minority class label}
        \STATE $O_i$ $\leftarrow$ $S_i$
    \ELSE 
        \STATE $O_i$ $\leftarrow$ $P_i$
    \ENDIF
    \end{algorithmic}
\end{algorithm}
\section{Experimental Results}
\label{sec:results}
The Ulm State of Mind Elderly (USOMS-e) database consists of 87 participants (55 f), who reported three narratives each. These are scored for valence and arousal, which are then grouped into three levels. For further details, see~\cite{compare2020}. 

\subsection{Experiments with Baseline Feature Sets}
Using the conventional train-to-development setting, we obtained the best development set arousal Unweighted Average Recall (UAR) of 44.4\% using PCA reduced auDeep-60 features~\cite{freitag2017audeep} with an SVM classifier that yielded a slightly lower test set UAR score of 42.7\%. It should be noted that the best test set arousal performance (UAR=50.4\%) is obtained using DeepSpectrum ResNet50~\cite{amiriparian2017snore} features, where the development set UAR was 35.0\%.  
Using the baseline feature sets, our best development set valence UAR=56.1\% was obtained with the combined BERT~\cite{devlin2018bert} feature dubbed `BLAtt' and German POS tags (BLAtt+POS) features modeled with a GBM classifier. This linguistic system yielded a test set score of 42.8\%. We attribute this performance mismatch partly to the model optimization/training procedure as discussed in Section~\ref{subsec:proptim}. Hereafter, we follow our pipeline using the proposed features with 4-Fold CV for model training. 

\subsection{Experiments with FV Representation}
\label{subsec:fv_exp}
We carried out FV feature extraction using a number of GMM components, $K_{GMM} \in$ \{16, 32, 64, 128\}. Although there was one annotation for each story, the challenge utterances were provided as chunks of 5 seconds, particularly for acoustic modeling. We hypothesized that summarizing the LLDs over each story and carrying out direct classification on story-wise FVs would yield a better performance than representing chunks and then carrying out decision fusion over each development/test set story. This hypothesis was tested on the training/development set using $K_{GMM}$=64. We observed that this approach boosted both arousal (from 42.7\%  to 48.2\%) and valence (from 45.8\% to 51.0\%) UAR performance on the development set. Thus, we conducted remaining experiments using story-wise FVs.

Noting a small number of story-wise instances (87 train + 87 dev = 174), we preferred smaller $K_{GMM}$ if a similar performance is obtained. Using FV representation with 16 GMM components, we obtained a 4-Fold CV overall UAR score of 48.7\% (KELM, linear kernel) for arousal and 52.0\% for valence (KPLS, linear kernel). We further reduced the dimensionality of FV using PCA and obtained the best UAR performance with 150 and 160 PCA dimensions for arousal (50.1\% with KELM, 45.1\% with WKELM) and valence (55.3\% with KPLS, 53.8\% with WKPLS), respectively. The corresponding weighted score fusion gave a 4-Fold CV UAR scores of 51.9\% and 58.2\% for arousal and valence, respectively. We used this system for our test set submission.

\subsection{Experiments with Linguistic Features}
\label{subsec:ling_exp}
Using the experimental setup defined in Section~\ref{subsec:proptim}, we evaluated the performance of different combinations of linguistic features, as well as combinations of the models with hard-label-based majority voting. The linguistic models were trained using SVMs with linear, sigmoid and radial basis function kernels. As can be seen in Table~\ref{table_ling}, each proposed linguistic approach performed significantly better than the baseline reported in the challenge paper for valence recognition~\cite{compare2020}. Moreover, a combination of these models at the decision level, namely the Ensemble Model, contributed to the overall performance. On the other hand, as expected, we obtained poor performance using linguistic features for arousal classification. 

Considering the best UAR performances of the baseline system for valence (56\% for development and 49\% for test) that uses the BERT model~\cite{devlin2018bert}, and those of our FastText features~\cite{bojanowski2017enriching} (4-fold CV UAR: 46.5\%), we observed that using only the complex contextual/semantic word embeddings to represent the story may be insufficient when the amount of labeled data is relatively small.
But enriching those features with the knowledge learned from some external resources (such as tonal dictionaries or sentiment analysis tools) that are trained on much larger data sets can greatly improve the performance. 

Except high-level polarity features, we conducted experiments for all models using both English translations and German transcripts. For FastText and TF-IDF models, using only English text gave a slightly better performance than using bilingual text. Thus, for these features, we reported the models that employ feature extraction from English text. Regarding dictionary based features, although the original text was in German, English dictionary-based features performed better than German ones, while fusion of the two followed by feature selection yielded the best performance. Brute-forcing different feature combinations resulted in the following set of optimal dictionary-based features: 2 from SentiWordNet (maximum positive and sum of negative scores) and 3 from SentiWS (minimum, maximum and number of negative scores). This small set of dictionary-based features outperformed the state-of-the-art BERT features on the development set by an absolute difference of 16.2\%. 
\begin{table}[ht]
\caption{4-Fold CV Average UAR (\%) of Linguistic Models. Ensemble Model: (1, 4, 5) for valence, and (1, 3, 4, 5) for arousal. } 
\label{table_ling}
\centering 
\begin{tabular}{l c c c c} 
\toprule 
Features & Dimens. & Valence  & Arousal\\ [0.5ex] 
\hline 
(1) TF-IDF features & 20337 & 52.3 & 33.8\\
(2) FastText features & 100 & 46.5 & 31.3\\
(3) Polarity features & 7 & 57.0 & 40.4\\
(4) FastText+Polarity features & 107 & 60.9 & 36.3 \\ 
(5) Dictionary-based features & 5 & 61.9 & 34.4\\ 
\textbf{Ensemble Model} & - &\textbf{62.3} & 38.1\\
\bottomrule 
\end{tabular}
\end{table}
\subsection{Challenge Test Set Submissions}
A total of five test set submissions were evaluated to prevent learning on the sequestered test set. A summary of the test performances of the proposed features are given in Table~\ref{table:test_results}. These results are in line with our hypotheses and show that linguistic models have a high generalization ability. System 3 arousal performance on the test set also shows that domain-aware rule based fusion of two prediction sets given in Algorithm~\ref{alg:algorithm_fuse2} dramatically improves the performance. 
Reaching an average test set UAR performance of 60.6\%, we outperform the challenge baseline relatively by 21.9\%.

\begin{table}[ht]
\caption{Test Set UAR (\%) Performances} 
\centering 
\begin{tabular}{c p{3.5cm} c c } 
\toprule 
System & Description & Valence & Arousal\\ [0.5ex] %
\hline 
System 0 & Challenge baseline & 49.0 & 50.4\\ \hline
System 1 & FV based Ensemble  & 44.3 & 48.8\\
System 2 & Linguistic Ensemble & 63.7 & 41.2 \\ 
System 3 & Valence: Same as System 2, Arousal: Rule based fusion System 0 + System 1  & \textbf{63.7} & \textbf{57.5}\\ 
\bottomrule 
\end{tabular}
\label{table:test_results} 
\end{table}
\section{Conclusions}
\label{sec:conc}

The proposed bi-modal elderly emotion recognition system shows outstanding performance on both arousal and valence recognition tasks with mean UAR=60.6\%, beating the baseline mean by an absolute difference of 10.9\%. Simple linguistic features proposed in this study dramatically outperform state-of-the-art BERT systems both in terms of accuracy and generalization capability that shows their advantage when using a small dataset. Moreover, they provide explainable results in contrast to ``black-box" approaches. Linguistic modeling on automatically translated English text shows better performance than the original, because of better resources. Similar results were reported for other languages
~\cite{halfon2016assessing}. Together with N-Fold CV experimental set-up and careful ensemble building strategy, the proposed system allows obtaining high reliability in terms of performance on a blind test set, which is not always possible with a traditional train/development split. Finally, it is shown that domain and confusion matrix awareness proves to be of considerable practical importance at the final stage of decision-level fusion. Scripts of this work can be found at {https://github.com/gizemsogancioglu/elderly-emotion-SC}.

\section{Acknowledgements}
\label{sec:ackn}
This research was supported by the Russian Science Foundation (project No. 18-11-00145). 
\bibliographystyle{IEEEtran}

\bibliography{mybib}


\end{document}